# Bayesian Error-Bars for Belief Net Inference

**Tim Van Allen**
digiMine, Inc.
10500 NE 8th St. Floor 13
Bellevue, WA 98004
timv@digimine.com

**Russell Greiner**
Dep't of Computing Science
University of Alberta
Edmonton, AB T6G 2H1 Canada
greiner@cs.ualberta.ca

**Peter Hooper**
Dep't of Mathematical Sciences
University of Alberta
Edmonton, AB T6G 2G1 Canada
hooper@stat.ualberta.ca

## Abstract

A Bayesian Belief Network (BN) is a model of a joint distribution over a finite set of variables, with a DAG structure to represent the immediate dependencies between the variables, and a set of parameters (aka CPTables) to represent the local conditional probabilities of a node, given each assignment to its parents. In many situations, the parameters are themselves treated as random variables — reflecting the uncertainty remaining after drawing on knowledge of domain experts and/or observing data generated by the network. A distribution over the CPtable parameters induces a distribution for the response the BN will return to any "What is $Pr\{H \mid E\}$?" query. This paper investigates the distribution of this response, shows that it is asymptotically normal, and derives expressions for its mean and asymptotic variance. We show that this computation has the same complexity as simply computing the (mean value of the) response — i.e., $O(n \exp(w))$, where $n$ is the number of variables and $w$ is the effective tree width. We also provide empirical evidence showing that the error-bars computed from our estimates are fairly accurate in practice, over a wide range of belief net structures and queries.

## 1 Introduction

Bayesian belief nets (BNs), which provide a succinct model of a joint probability distribution, are used in an ever increasing range of applications [Hec95]. Belief nets are typically built by first finding an appropriate structure (either by interviewing an expert, or by selecting a good model from training data), then using a training sample to fill in the parameters [Hec98]. The resulting belief net is then used to answer questions, e.g., compute the conditional probability $Pr\{\text{Cancer=true} \mid \text{Smoke=true, Gender=male}\}$. These values, referred to as query responses, clearly depend on the training sample used to instantiate the parameter values — i.e., different training samples will produce different parameters and hence different responses.

This paper investigates how sampling variability in the training data is related to uncertainty about a query response. We follow the Bayesian paradigm, where uncertainty is quantified in terms of random variation, and we present a technique for computing Bayesian credible intervals (aka "error-bars") for query responses. Our algorithm takes as inputs a belief net structure (which we assume is correct — i.e., an accurate $I$-map of true distribution [Pea88]); a data sample generated from the true belief net distribution; and a specific query of the form "What is $Q \equiv Pr\{H = h \mid E = e\}$?". After determining the conditional (posterior) distribution of the belief net parameters given the sample, the algorithm produces an estimate (posterior mean value) of $Q$: e.g., estimate $Q$ to be 0.3. To quantify uncertainty about this estimate, the algorithm computes an approximate posterior variance for $Q$ and uses this variance to construct error-bars (a Bayesian credible interval) for $Q$; e.g., assert that $Q$ is in the interval $0.3 \pm 0.1$ with 90% probability.

There are several obvious applications for these error-bars. First, error-bars can help a user make decisions, especially in safety-critical situations — e.g., take action if we are 99% sure that $Q \equiv Pr\{H = h \mid E = e\}$ is on one side of a decision boundary. Second, error-bars can suggest that more training data is needed before the system can make appropriate guarantees about the answers to certain queries. This information is especially valuable when additional training data, while available, is costly, and its acquisition needs to be justified. Similarly, the user might decide that more evidence is needed about a specific instance, before he can render a meaningful decision. Finally, if an expert is available and able to provide "correct answers" to some specific questions, error-bars can be used to validate the given belief net structure. E.g., if the expert claims that $Q = 0.5$ but our algorithm asserts that $Q$ is in the interval $0.30 \pm 0.04$ with 99.9% probability, then we may question whether the structure provided is correct (assuming we believe the expert). By contrast, we might not question this structure if our algorithm instead asserted that $Q$ is in the interval $0.30 \pm 0.25$ with 99.9% probability.

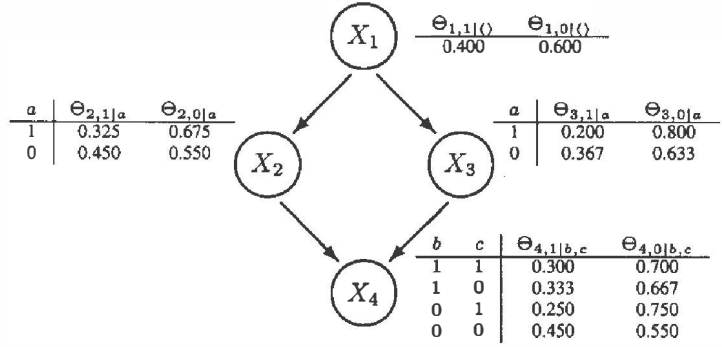

Figure 1: Simple Example: Diamond Graph

Section 2 provides background results and notation concerning belief nets and Dirichlet distributions for belief net parameters. Section 3 presents the theoretical results underlying our error-bars: a derivation of an approximate posterior variance for a query probability $Q$, and a proof that the posterior distribution of $Q$ is asymptotically normal. Computational issues related to calculation of the variance are briefly discussed. Section 4 presents the results of an empirical study using Monte Carlo simulations to validate our error-bar methodology over a wide range of belief net structures and queries. Section 5 briefly surveys related work, placing our results in context.

## 2 Belief nets and Dirichlet distributions

We encode the joint distribution of a vector of discrete random variables $X = \langle X_v \rangle_{v \in \mathcal{V}}$ as a belief net (aka Bayesian network, probability net). A belief net $\langle \mathcal{V}, \mathcal{A}, \Theta \rangle$ is a directed acyclic graph whose nodes $\mathcal{V}$ index the random variables and whose arcs $\mathcal{A}$ represent dependencies. Let $\text{Pa}(v) \subset \mathcal{V}$ be the immediate parents of node $v$, and let $F_v = \langle X_w \rangle_{w \in \text{Pa}(v)}$ be the corresponding vector of parent variables. In a belief net, a variable $X_v$ is independent of its nondescendents, given $F_v$. The elements of the vector $\Theta$ are the CPtable entries

$$\Theta_{v,x|f} = Pr\{X_v = x \mid F_v = f, \Theta\}.$$

Let $\mathcal{X}_v$ and $\mathcal{F}_v = \prod_{w \in \text{Pa}(v)} \mathcal{X}_w$ be the domains of $X_v$ and $F_v$. We assume that the domains are finite. The CPtable for $X_v$ contains $|\mathcal{X}_v| \times |\mathcal{F}_v|$ entries $\Theta_{v,x|f}$.

Figure 1 provides a simple example of a belief network with specific CPtable entries. Here $X_1$ has no parents, so we write $F_1 = \langle \rangle$. We have $F_2 = \langle X_1 \rangle$, $F_3 = \langle X_1 \rangle$, $F_4 = \langle X_2, X_3 \rangle$; and for each value $a, b, c, d$, we have $\Theta_{1,a|\langle\rangle} = Pr\{X_1 = a \mid \Theta\}$, $\Theta_{2,b|a} = Pr\{X_2 = b \mid X_1 = a, \Theta\}$, and $\Theta_{4,d|b,c} = Pr\{X_4 = d \mid X_2 = b, X_3 = c, \Theta\}$.

(Hence, using Figure 1, we have $\Theta_{1,1|\langle\rangle} = 0.4$.) Note that the values in each row add up to 1. In general, the variables need not be binary, but can have larger (finite) domains.

The CPtable entries are estimated using training data and (possibly) expert opinion. The latter information is incorporated using the Bayesian paradigm, where $\Theta$ is modeled as a random variable and expert opinion is expressed through an a priori distribution for $\Theta$. We adopt independent Dirichlet priors[1] for the various CPtable rows. Specifically, let $\Theta_{v|f} = \langle \Theta_{v,x|f} \rangle_{x \in \mathcal{X}_v}$ denote the CPtable row for $F_v = f$ — e.g., $\Theta_{4|\langle 1,0 \rangle} = \langle \Theta_{4,1|\langle 1,0\rangle}, \Theta_{4,0|\langle 1,0\rangle}\rangle$ denotes the entries for the $X_4$ variable associated with the parental assignment $X_2 = 1$ and $X_3 = 0$. We assume that, before observing the training data, the $\Theta_{v|f}$ are independent "$Dir(\alpha^*_{v,x|f}, x \in \mathcal{X}_v)$" random vectors, where $\alpha^*_{v,x|f} > 0$. An absence of expert opinion is often expressed by setting $\alpha^*_{v,x|f} = 1$ for all $(v, x, f)$ — e.g., $\Theta_{4|\langle 1,0\rangle} \sim Dir(1, 1)$ — which yields a uniform (flat) prior. Stronger opinion is expressed through larger values of $\alpha^*_{v,x|f}$. Expressions for the mean and variance of a Dirichlet distribution are given below.

Now suppose that the training data consist of $m$ independent replicates of vectors $X$, generated using the given structure and a fixed set of CPtable entries $\Theta$. Let $m_{v,x|f}$ denote the number of cases in the training set with $(X_v, F_v) = (x, f)$. Under the posterior distribution (the conditional distribution given the training data), the $\Theta_{v|f}$ are independent $Dir(\alpha_{v,x|f}, x \in \mathcal{X}_v)$ random vectors, with $\alpha_{v,x|f} = \alpha^*_{v,x|f} + m_{v,x|f}$ [BFH95]. This posterior distribution underlies our derivation of Bayesian credible intervals. Several properties of the Dirichlet distribution will be needed.

Setting $\alpha_{v,\cdot|f} = \sum_{x \in \mathcal{X}_v} \alpha_{v,x|f}$, the posterior means and (co)variances for CPtable entries are [BFH95]:

$$E\{\Theta_{v,x|f}\} = \mu_{v,x|f} = \frac{\alpha_{v,x|f}}{\alpha_{v,\cdot|f}} \quad (1)$$

$$\text{Cov}\{\Theta_{v,x|f}, \Theta_{v,y|f}\} = \frac{\mu_{v,x|f}(\delta_{xy} - \mu_{v,y|f})}{\alpha_{v,\cdot|f} + 1} \quad (2)$$

---

[1] Readers unfamiliar with these assumptions, or with Dirichlet distributions, are referred to [Hec98]. Note that a Dirichlet distribution over a *binary* variable is a Beta distribution.

where $\delta_{xy} = 1$ if $x = y$ and $\delta_{xy} = 0$ otherwise. The random vectors $\Theta_{v|f}$ are asymptotically normal, in the limit as $\min_x \alpha_{v,x|f} \to \infty$ [Aki96]. More precisely, the normalized variables $\sqrt{\alpha_{v,\cdot|f}}(\Theta_{v,x|f} - \mu_{v,x|f})$ converge in distribution to jointly normal random variables with mean zero and covariances $\mu_{v,x|f}(\delta_{xy} - \mu_{v,y|f})$. This asymptotic framework is applicable as the amount of training data increases ($m \to \infty$) provided all of the CPtable entries $\Theta_{v,x|f}$ are positive. This condition occurs with probability one under a Dirichlet prior.

## 3 Bayesian Credible Intervals for Query Responses

It is well-known that the CPtable entries determine the joint distribution of $X$: $Pr\{X_v = x_v, v \in \mathcal{V} | \Theta\} = \prod_{v \in \mathcal{V}} \Theta_{v,x_v|f_v}$, where $(f_v)_{v \in \mathcal{V}}$ is determined by $(x_v)_{v \in \mathcal{V}}$; see [Pea88]. Users are typically interested in one or more specific "queries" asked of this joint distribution, where a query is expressed as a conditional probability of the form

$$Q = q(\Theta) = Pr\{H = h | E = e, \Theta\}, \quad (3)$$

where $H$ and $E$ are subvectors of $X$, and $h$ and $e$ are legal assignments to these subvectors. Note also the dependency on $\Theta$.

In our Bayesian context, $Q$ is a random variable with a (theoretically) known distribution determined by the posterior distribution of $\Theta$. For a point estimate of $Q$, one may use the posterior mean $\mu_Q \equiv E\{Q\}$. This value can be calculated using the identity [CH92]:

$$E\{q(\Theta)\} = q(E\{\Theta\}).$$

Set $\mu = E\{\Theta\}$ where the components $\mu_{v,x|f}$ of $\mu$ are defined by Equation 1.

While a point estimate $\mu_Q = q(\mu)$ can be useful, one often requires some information concerning the potential error in the estimate. In the Bayesian context, this can be achieved by plotting the posterior distribution of $Q$. Alternatively, one may construct a $100(1 - \delta)\%$ credible interval for $Q$; i.e., an interval $(L, U)$ defined so that $Pr\{L \leq Q \leq U\} = 1 - \delta$. Exact calculations are typically not analytically tractable, but simple approximations are available. We will show that the distribution of $Q$ is approximately normal, and derive an approximation $\tilde{\sigma}_Q$ for the standard deviation of $Q$. We then propose the following interval as an approximate $100(1 - \delta)\%$ credible interval:

$$\mu_Q \pm z_{\delta/2}\, \tilde{\sigma}_Q, \quad (4)$$

where $z_{\delta/2} = \Phi^{-1}(1 - \delta/2)$ is the upper $\delta/2$ value of the standard normal distribution.

Our derivation is based on a first-order Taylor expansion of $q(\Theta)$ about $q(\mu)$. Some notation is needed to express the partial derivatives. Let $p_v(h, x, f | e)$ denote the probability

$$Pr\{H = h, X_v = x, F_v = f | E = e, \Theta = \mu\},$$

and let $p_v(x, f | e)$, $p_v(h, f | e)$, $p_v(f | e)$, and $p(h | e)$ be defined in a similar manner. Note that the subscript $v$ is needed to identify the node when $X_v$ or $F_v$ is involved, and all probabilities are evaluated at $\Theta = \mu$. Let $q'_{v,x|f}$ denote the partial derivative $\partial q(\theta)/\partial \theta_{v,x|f}$ evaluated at $\theta = \mu$. We will use the following identity, derived by [GGS97, Dar00]:

$$q'_{v,x|f} = \frac{p_v(h, x, f | e) - p(h | e)\, p_v(x, f | e)}{\mu_{v,x|f}}. \quad (5)$$

We now derive an expression for $\tilde{\sigma}_Q^2$, and demonstrate asymptotic validity of the credible interval (Equation 4) given a sufficiently large training sample.

**Theorem 1** *We assume that $\Theta$ is a random vector with posterior Dirichlet distribution described in Section 2, and approximate the variance of $Q = q(\Theta)$ by*

$$\tilde{\sigma}_Q^2 = \sum_{v \in \mathcal{V}} \sum_{f \in \mathcal{F}_v} (A_{vf} - B_{vf})/(\alpha_{v,\cdot|f} + 1), \quad (6)$$

*where*

$$A_{vf} = \sum_{x \in \mathcal{X}_v} \frac{\{p_v(h, x, f | e) - p(h | e)\, p_v(x, f | e)\}^2}{\mu_{v,x|f}},$$

$$B_{vf} = \{p_v(h, f | e) - p(h | e)\, p_v(f | e)\}^2.$$

*Consider an asymptotic framework where the posterior means $\mu_{v,x|f}$ are fixed, positive values, and $\min\{\alpha_{v,x|f}\} \to \infty$. Then the random variable $(Q - \mu_Q)/\tilde{\sigma}_Q$ converges in distribution to the standard normal distribution.*

**Proof.** *Our proof uses the Delta method [BFH95]. Consider the Taylor expansion*

$$q(\Theta) = q(\mu) + D + R,$$

*where*

$$D = \sum_{v \in \mathcal{V}} \sum_{f \in \mathcal{F}_v} \sum_{x \in \mathcal{X}_v} q'_{v,x|f}(\Theta_{v,x|f} - \mu_{v,x|f}). \quad (7)$$

*and the remainder term $R$ can be expressed in terms of the matrix of second derivatives of $q(\Theta)$ evaluated at a point $\bar{\Theta}$ between $\Theta$ and $\mu$. Since the variances for $\Theta_{v,x|f}$ in Equation 2 are of order $1/\alpha_{v,x|f} \to 0$, and since the second derivatives remain bounded in a neighbourhood of $\mu$, the remainder $R$ is asymptotically negligible compared with $D$.*

*We define $\tilde{\sigma}_Q^2$ to be the variance of $D$ (Equation 7). As the CPtable rows $\Theta_{v|f}$ are statistically independent, but* *entries within a row are correlated, the variance of $D$ can be expressed as*

$$\sum_{v \in \mathcal{V}} \sum_{f \in \mathcal{F}_v} \sum_{x \in \mathcal{X}_v} \sum_{y \in \mathcal{X}_v} q'_{v,x|f} \, q'_{v,y|f} \, \text{Cov}\{\Theta_{v,x|f}, \Theta_{v,y|f}\}.$$

*After substituting Equation 2 for the covariances and simplifying, we obtain Equation 6 with*

$$A_{vf} = \sum_{x \in \mathcal{X}_v} (q'_{v,x|f})^2 \mu_{x,v|f},$$

$$B_{vf} = \left( \sum_{x \in \mathcal{X}_v} q'_{v,x|f} \, \mu_{x,v|f} \right)^2.$$

*A substitution of Equation 5 then yields the equivalent expressions for $A_{vf}$ and $B_{vf}$ within Equation 6.*

*We observe that $D/\tilde{\sigma}_Q$ is a random variable with mean 0 and variance 1. It remains to show that $D/\tilde{\sigma}_Q$ is asymptotically normal. This result follows from the asymptotic multivariate normality of the components of $\Theta$ (after suitable standardization — see Section 2), and the fact that $D$ is a linear function of $\Theta$.* □

**Degenerate Case:** There are exceptional situations where the posterior distribution of $q(\Theta)$ is analytically tractable and exact credible intervals are available. In the degenerate situation where the network structure has arcs connecting *all* pairs of nodes (and hence imposes no assumptions about conditional independence), the assumption of independent Dirichlet distributions for CPtable rows is equivalent to an assumption of a single Dirichlet distribution over unconditional probabilities $Pr\{X_v = x_v, v \in \mathcal{V}\}$. It is then straightforward to derive the distribution of the query probability using properties of the Dirichlet distribution; see [Mus93].[2] Note that this exact approach is *not* correct in general — i.e., it does not hold for networks with non-trivial structure.[3]

**Computational Issues:** The computational problem of computing $\mu_Q = q(\mu)$ is known to be NP-hard [Coo90]; when all variables $X_v$ are binary, the most effective exact algorithms require time $O(n2^w)$, where $n = |\mathcal{V}|$ is the number of nodes and $w$ is the induced tree width of the graph [Dec98, LS99]. The variance $\tilde{\sigma}_Q$ can also be computed in time $O(n2^w)$. This result follows from the existence of algorithms that can compute all of the derivatives $q'_{v,x|f}$ in time $O(n2^w)$; see [Dar00]. Given these derivatives, the summations in Equation 6 can be performed with one additional pass over the values, of time $O(n)$.

---

[2] Assuming a uniform prior and a sample of size $m$, we can compute the posterior variance of $Pr\{H \mid E\}$ as $\hat{P}\{H|E\} \times (1 - \hat{P}\{H|E\})/((m \times \hat{P}\{E\})+3)$, where $\hat{P}(x)$ is the expected value of $x$, wrt the given belief net.

[3] This follows from a dimensionality argument: in a non-trivial structure, the $2^n$-dimensional vector of unconditional probabilities is constrained to lie in a lower-dimensional submanifold of the $2^n - 1$-dimensional simplex. This cannot be represented by a single Dirichlet distribution because, wp1, the constraints would not be satisfied.

Table 1: Gold Standard for Validity Estimates

| $\delta$ | Mean | Std.Dev. |
|---|---|---|
| 10% | 2.38 | 1.86 |
| 20% | 3.15 | 2.41 |
| 30% | 3.63 | 2.79 |
| 40% | 3.88 | 2.96 |

The extended paper [VGH01] describes an algorithm for computing $\tilde{\sigma}_Q$. The main challenge, computing all of the derivatives $q'_{v,x|f}$, is accomplished by "back propagating" intermediate results obtained by the Bucket Elimination [Dec98] algorithm.

[VGH01] also provides additional comments on the proper interpretation and application of this theorem.

## 4 Empirical Study

Theorem 1 proves that the interval $\mu_Q \pm z_{\delta/2} \tilde{\sigma}_Q$ is asymptotically valid. More precisely, let

$$\Delta = Pr\{|Q - \mu_Q| > z_{\delta/2} \tilde{\sigma}_Q\} \quad (8)$$

be the probability that the query response $Q$ falls outside of the credible interval, based on our $\tilde{\sigma}_Q$ estimate of standard deviation, Equation 6. The values $1 - \delta$ and $1 - \Delta$ are the *nominal* and *actual* coverage probabilities for the credible interval. The value $\Delta$ is a function of $\delta$, the graph $\langle \mathcal{V}, \mathcal{A} \rangle$, the query $q$, and the posterior distribution of $\Theta$. The posterior distribution depends on the prior distribution and the training sample. Thus $\Delta$ typically varies from one application to the next. While Theorem 1 implies that $\Delta \approx \delta$ when the training sample is sufficiently large, it does not tell us whether this approximation is valid in practice, particularly for small samples. In general, the validity of the approximation depends on all of the factors determining $\Delta$. We carried out a number of experiments to assess how these factors affect validity.

Given a fixed set of factors, we estimate the corresponding $\Delta$ by a simple Monte Carlo strategy. Using the (fixed) posterior distribution of $\Theta$, calculate $\mu_Q$ and $\tilde{\sigma}_Q$. Simulate $r$ replicates $\Theta_i$ from the posterior distribution, calculate $Q_i = q(\Theta_i)$, then let $\hat{\Delta}$ be the proportion of the $\{Q_i\}$ with $|Q_i - \mu_Q| > z_{\delta/2} \tilde{\sigma}_Q$. In our experiments, each $\hat{\Delta}$ was based on $r = 100$ replicates.

To quantify the validity of the approximation $\Delta \approx \delta$, we employ average absolute differences:

$$\text{validity estimate} = \text{average } |\hat{\Delta} - \delta|. \quad (9)$$

The absolute differences are averaged as we vary one or more of the the factors determining $\Delta$. The validity estimates are presented as percentages in our tables. When

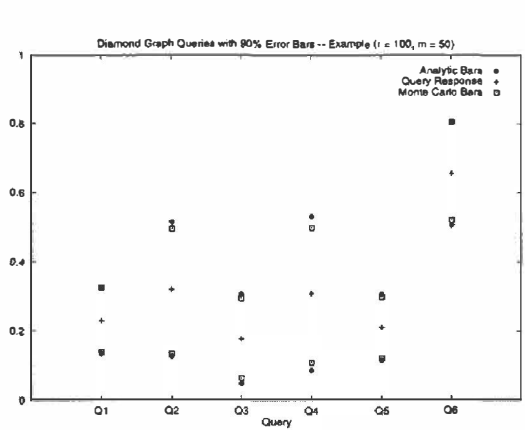

Figure 2: (A) Examples of Error Bars;

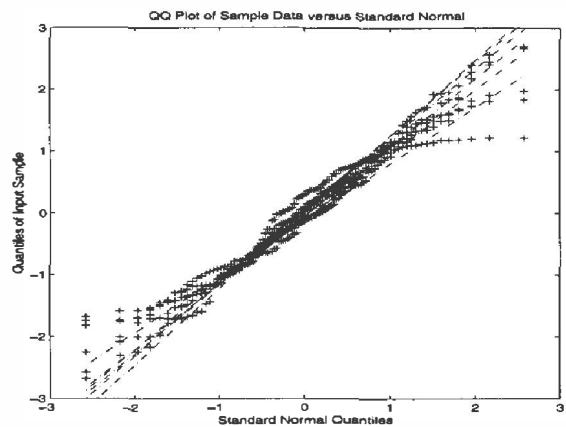

(B) QQ-plot showing relation to Normal

viewing these values, it is helpful to have a gold standard for comparison. Consider the validity estimate $|\hat{\Delta} - \delta|$ for a single $\Delta$. The minimum expected value is obtained when $\Delta = \delta$; i.e., when $100\hat{\Delta}$ has the Binomial$(100, \delta)$ distribution. Table 1 presents means and standard deviations under these ideal circumstances. Now suppose a validity estimate is obtained by averaging $k$ independent terms. Its standard deviation is typically greater than the value Std.Dev.$/\sqrt{k}$ suggested by Table 1 because there is usually variation in the underlying $\Delta$ values.

### 4.1 Results for the Diamond Graph

We studied the following inferential patterns in the diamond graph (Figure 1):

$Q_1 = \Pr\{X_1 = 1 \,|\Theta\} = \Theta_{1,1|\langle\rangle}$

$Q_2 = \Pr\{X_1 = 1 \,|X_2 = 1, \Theta\} = \frac{\Theta_{2,1|1} \, \Theta_{1,1|\langle\rangle}}{\sum_a \Theta_{2,1|a} \, \Theta_{1,a|\langle\rangle}}$

$Q_3 = \Pr\{X_1 = 1 \,|X_2 X_3 = 1, \Theta\} = \frac{\Theta_{2,1|1} \, \Theta_{3,1|1} \, \Theta_{1,1|\langle\rangle}}{\sum_a \Theta_{2,1|a} \, \Theta_{3,1|a} \, \Theta_{1,a|\langle\rangle}}$

$Q_4 = \Pr\{X_2 X_3 = 1 \,|X_1 = 1, \Theta\} = \Theta_{2,1|1} \, \Theta_{3,1|1}$

$Q_5 = \Pr\{X_1 = 1 \,|X_4 = 1, \Theta\}$
$= \frac{\sum_{b,c} \Theta_{4,1|b,c} \, \Theta_{2,b|1} \, \Theta_{3,c|1} \, \Theta_{1,1|\langle\rangle}}{\sum_{a,b,c} \Theta_{4,1|b,c} \, \Theta_{2,b|a} \, \Theta_{3,c|a} \, \Theta_{1,a|\langle\rangle}}$

$Q_6 = \Pr\{X_4 = 1 \,|X_1 = 1, \Theta\} = \sum_{b,c} \Theta_{4,1|b,c} \, \Theta_{2,b|1} \, \Theta_{3,c|1}$

The six queries cover a range of different inferential patterns. The first is basically a "sanity check", as it is a trivial inference; the fourth is also straightforward, although it does involve a multiplication. The sixth is slightly more complex, but it is still only a summation of a set of products. The remaining queries involve divisions of increasingly complicated expressions.

For each $m \in \{10, 20, 30, 40\}$, we carried out 30 trials of the form: (1) generate $\Theta$ from a uniform Dirichlet prior distribution, (2) generate a training sample of size $m$ based on $\Theta$ and use the result to obtain a posterior distribution, (3) generate 100 Monte Carlo replicates from the posterior distribution and use these to obtain an estimate $\hat{\Delta}$ for each pair $(Q, \delta)$, for $Q \in \{Q_1, \ldots, Q_6\}$ and $\delta \in \{10\%, 20\%, 30\%, 40\%\}$. The resulting validity estimates are listed in Table 2. Each cell in the table is an average of 30 values.

Figure 2(A) shows the error-bars returned by our approximation, and also the Monte Carlo system, on a random network posterior, for the error-bars for 90% credible intervals. We see the two methods give similar answers.

Figure 2(B) uses a quantile-quantile (QQ) plot to address the validity of the normality assumption, *independently* of the linear approximation. Each "line" in this figure corresponds to z-scores of the 100 query responses generated by our Monte Carlo simulation, plotted against standard normal quantiles. This figure shows six such lines, each corresponding to a single query in $\{Q_1, \ldots, Q_6\}$, given a sample of size $m = 10$. A straight-line would correspond to data produced by a "perfect" normal distribution; we see each dataset is close. (Of course, this is only suggestive; the real proof comes first from Theorem 1, and then from the data (e.g., Table 2) which demonstrates that our approach, which assumes normality, produces reasonable results.)

### 4.2 Results for Alarm Network

The Alarm network [BSCC89] is a benchmark network based on a medical diagnosis domain, commonly used in belief network studies. The network variables are all discrete, but many range over 3 or more values. The network includes a CPtable for each node; i.e., a particular $\theta$ is specified.

Table 3 summarizes the results for experiments on the Alarm network, where we varied both $\delta$ and $m$. For each $m$, we generated a single random sample of size $m$ from $\theta$, and used this to determine a posterior distribution (assuming a uniform prior). Validity estimates were obtained by averaging over randomly chosen queries. The queries $\Pr\{H = h \,|E = e, \theta\}$ were chosen by determining an assignment $H = h$ to one randomly chosen query variable, and assignments $E = e$ to five randomly chosen evidence variables. (Here, we used [HC91] to determine which vari-ables could be query as opposed to evidence variables.) Some or all of the evidence variables might have had no effect on the query variable, others might have had a profound effect. Each cell in Table 3 represents an average from 100 queries on a single posterior distribution.

Table 2: Results for Diamond Graph

| $m$ | $Q_1$ | $Q_2$ | $Q_3$ | $Q_4$ | $Q_5$ | $Q_6$ |
|---|---|---|---|---|---|---|
| | | | $\delta = 10\%$ | | | |
| 10 | 2.37 | 2.77 | 3.10 | 3.27 | 2.20 | 3.93 |
| 20 | 2.67 | 3.33 | 2.50 | 3.37 | 2.60 | 3.50 |
| 30 | 2.60 | 3.03 | 2.40 | 3.13 | 2.90 | 3.70 |
| 40 | 2.60 | 2.97 | 3.10 | 2.00 | 2.77 | 2.90 |
| | | | $\delta = 20\%$ | | | |
| 10 | 3.50 | 5.00 | 4.70 | 4.87 | 3.43 | 5.57 |
| 20 | 4.60 | 6.27 | 5.13 | 7.03 | 4.03 | 4.53 |
| 30 | 2.90 | 5.07 | 4.43 | 5.97 | 3.97 | 4.87 |
| 40 | 4.07 | 5.27 | 4.93 | 4.93 | 4.03 | 3.87 |
| | | | $\delta = 30\%$ | | | |
| 10 | 3.63 | 5.63 | 6.10 | 7.23 | 5.13 | 6.27 |
| 20 | 4.70 | 7.20 | 6.83 | 11.13 | 5.30 | 6.03 |
| 30 | 3.70 | 5.80 | 5.47 | 7.20 | 3.50 | 5.30 |
| 40 | 5.13 | 6.97 | 6.03 | 6.63 | 4.27 | 4.27 |
| | | | $\delta = 40\%$ | | | |
| 10 | 4.33 | 3.97 | 5.33 | 7.53 | 4.40 | 6.63 |
| 20 | 4.73 | 5.20 | 5.73 | 9.27 | 5.20 | 5.97 |
| 30 | 3.33 | 4.50 | 5.00 | 6.47 | 4.27 | 4.97 |
| 40 | 3.90 | 4.90 | 5.97 | 6.73 | 4.43 | 4.47 |

Table 3: Results for Alarm Network

| | $\delta$ | | | |
|---|---|---|---|---|
| $m$ | 10% | 20% | 30% | 40% |
| 50 | 2.47 | 4.37 | 4.48 | 4.07 |
| 100 | 2.66 | 4.95 | 5.97 | 4.87 |
| 150 | 3.04 | 5.35 | 6.45 | 5.66 |
| 200 | 2.65 | 4.80 | 5.43 | 5.42 |

### 4.3 Results for Random Networks

Although random networks tend not to reflect typical (or natural) domains, they complement more focussed studies by exposing methods to a wide range of inputs and help to support claims of generality. We carried out experiments on networks with 10 binary variables and 20 links, generating gold models from a uniform prior distribution on $\Theta$, and generating random queries of various types. Here we used sample size $m = 100$ throughout, and varied the type of query. Table 4 displays the results of our experiments. Each query was of the form $\Pr\{H = h \,|E = e, \Theta\}$, with varying dimensionalities for $E$ and $H$. Let $\#E$ and $\#H$ denote the number of variables comprising $E$ and $H$, respectively. Each cell of Table 4 is based on 100 trials: 10 queries on 10 networks, with both structure and posterior generated randomly.

Table 4: Results for Random Networks

| | $\#E$ | | | | |
|---|---|---|---|---|---|
| $\#H$ | 1 | 2 | 3 | 4 | 5 |
| | | | $\delta = 10\%$ | | |
| 1 | 2.16 | 2.72 | 2.84 | 3.20 | 3.00 |
| 2 | 2.59 | 2.50 | 2.93 | 3.13 | 2.45 |
| 3 | 2.49 | 2.59 | 2.62 | 2.38 | 2.56 |
| 4 | 2.57 | 2.26 | 2.84 | 2.58 | 2.88 |
| 5 | 2.72 | 2.53 | 2.61 | 3.05 | 2.79 |
| | | | $\delta = 20\%$ | | |
| 1 | 3.06 | 4.16 | 4.41 | 4.19 | 4.39 |
| 2 | 3.60 | 3.63 | 4.38 | 4.71 | 4.96 |
| 3 | 3.50 | 4.16 | 4.56 | 5.07 | 5.78 |
| 4 | 4.12 | 4.46 | 5.64 | 6.31 | 7.14 |
| 5 | 4.67 | 5.76 | 6.63 | 7.85 | 8.17 |
| | | | $\delta = 30\%$ | | |
| 1 | 4.17 | 4.78 | 5.11 | 4.75 | 5.97 |
| 2 | 4.46 | 4.28 | 5.02 | 5.13 | 6.33 |
| 3 | 4.12 | 4.74 | 5.14 | 6.39 | 7.25 |
| 4 | 4.98 | 5.61 | 7.43 | 8.71 | 10.81 |
| 5 | 5.69 | 7.63 | 9.45 | 12.51 | 13.99 |
| | | | $\delta = 40\%$ | | |
| 1 | 4.20 | 4.90 | 5.00 | 4.56 | 5.62 |
| 2 | 4.43 | 4.24 | 4.81 | 5.44 | 6.43 |
| 3 | 4.11 | 4.62 | 4.80 | 5.95 | 7.12 |
| 4 | 5.15 | 4.52 | 6.47 | 8.36 | 10.99 |
| 5 | 4.95 | 7.14 | 8.95 | 13.05 | 16.15 |

### 4.4 Discussion

Our hypothesis was that our Bayesian error-bars algorithm would be accurate for essentially all cases. We tried to falsify our hypothesis by varying the following experimental factors:

- Network structure $\langle \mathcal{V}, \mathcal{A} \rangle$
- Credibility level $1 - \delta$
- Query type (Diamond network, Alarm)
- Number of evidence variables (Random networks)
- Number of query variables (Random networks)

In no case did we observe a result where $average|\hat{\Delta} - \delta|$ exceeded 20%. In most cases, the validity estimate was less than $\delta/3$. As noted in Table 1, even if our error-bars were exact, we would still get positive validity estimates due to the variance in $\hat{\Delta}$ about $\Delta$. We therefore believe that these results comfortably bound the expected error of our method under the experimental conditions. None of the factors that we manipulated had a profound effect. The strongest effect, observed in Table 4, was that increasing the number of variables assigned in a query tended to increase the error $|\Delta - \delta|$; see also [Kle96]. One possible explanation is that, as $\#E$ and $\#H$ increase, the query function $q$ tends to become more complex, and the local linear approximation of $q$ becomes less reliable. Another possibility is that the query probability $Q$ tends to become very small, making the normal approximation less accurate. Further experiments could address this issue.

We found these results very encouraging. Our method appears to give reasonable error-bars for a wide range of queries and network types. This makes the technique a promising addition to the array of data-analysis tools related to belief networks, especially as the algorithm is reasonably efficient, (only) roughly doubling the computation time per inference. While there may be pathological cases where our method will not give reasonable results — perhaps because the local linear approximation and the asymptotic normality are far off the truth — we did not find such cases in our experiments.

**Other Experiments:** We also ran a number of other experiments. One set computed the $average\{\hat{\Delta} - \delta\}$ scores in each situation, to determine if there was any systematic bias. (Note this score differs from Equation 9 by *not* taking absolute values.) We found that our bounds were typically a bit too wide for most queries — i.e., we often found the $1 - \delta$-interval included slightly more than $1 - \delta$ of the cases. We are currently investigating this, to see if there are straight-forward refinements we can incorporate.

We also computed error-bars based on the (incorrect!) "complete structure" assumption, which implies the response will have a simple Dirichlet distribution; see Footnote 2. We found that, as anticipated, the approach described in this paper, using Equation 6, consistently outperformed that case, in that our approach was consistently closer to the Monte Carlo estimates.

[VGH01] discusses these results in detail. It also investigates techniques for dealing with extreme values, where the normal distribution may be sub-optimal.

## 5 Related Work

Our results provide a way to compute the variance of a BN's response to a query, which depends on the posterior distribution over the space of CPtable entries, based on a data sample. This is done using the "Delta method" [BFH95]: first determine the variance of each CPtable row, then propagate this variance using a sensitivity analysis (*i.e.*, the partial derivatives); see Equation 6. Kleiter [Kle96] performs a similar computation; parts of his analysis are more general, in that he considers incomplete data. However, he does not (1) discuss how to deal with general graphical structures, (2) show how to deal with the correlations encountered with general Dirichlet distributions, nor (3) provide an efficient way to compute this information. Moreover, our empirical data provide additional evidence that the approximations inherent in this approach are appropriate, even for small sample sizes.

Several other researchers also consider the *posterior distribution over CPtables*, but for different purposes. For example, Cooper and Herskovits [CH92] use it to compute the expected response to a query; by contrast, we also approximate the posterior *variance* in that response. Similarly, while many BN-learning algorithms compute the posterior distribution over CPtables [Hec98], most of these systems seek a single set of CPtable entries that maximize the likelihood, which again is different from our task; *e.g.*, their task is not relative to a specific query (but see [GGS97]).

Many other projects consider *sensitivity analyses*, providing mechanisms for propagating ranges of CPtable values to produce a range in the response; cf., [BKRK97, Las95, CNKE93, Dar00]. While these papers assume the user is explicitly specifying the range of a local CPtable value, our work considers the source of these variances based on a data sample. This also means our system must propagate *all* of the "ranges"; most other analyses consider only propagating a single range. The [Dar00] system is an exception, as it can simultaneously produce all of the derivatives. However, Darwiche does not consider our error-bar application, and so does not include the additional optimizations we could incorporate.

Excluding the [Dar00] result, none of the other projects provides an efficient way to compute that information. Also, some of those other papers focus on properties of this derivative — e.g., when it is 0 for some specific CPtable entry. Note that this information falls out immediately from our expression (Equation 6). Finally, our analysis holds for arbitrary structures; by contrast some other results (e.g., [CNKE93]) deal only with singly connected networks (trees).

Lastly, our analysis also connects to work on abstractions, which also involves determining how influential a CPtable entry is, with respect to a query, towards deciding whether to include a specific node or arc [GDS01]. Their goal is typically computational efficiency in computing that response. By contrast, our focus is in computing the error-bars around the response, independent of the time required to determine that result.

## 6 Conclusion

**Further Extensions:** Our current system has been implemented, and works very effectively. There are several obvious ways to extend it. One set of extensions correspond to discharging assumptions underlying Theorem 1: computing error bars when the data was used to learn the *structure*, as well as the parameters; dealing with parameters that are drawn from a distribution other than independent Dirichlets, perhaps even variables that have *continuous* domains; dealing with a training sample whose instances are not completely specified. Our work deals with fully-parameterized CP *tables*. It would be interesting to investigate techniques capable of dealing with CPtables represented as, say, decision tree functions [BFGK96], etc.

**Contributions:** Many real-world systems work by reasoning probabilistically, based on a given belief net model. When knowledge concerning model parameters is conditioned on a random training sample, it is useful to view the parameters as random variables; this characterizes our uncertainty concerning the responses generated to specific queries in terms of random variation. Bayesian error-bars provide a useful summary of our current knowledge about questions of interest, and so provide valuable guidance for decision-making or learning.

This paper addresses the challenge of computing the error-bars around a belief net's response to a query, from a Bayesian perspective. We first motivated and formally defined this task — finding the $100(1 - \delta)\%$ credible interval for a query response with respect to its posterior distribution, conditioned on a training sample. We then investigated an application of the "Delta method" to derive these intervals. This required determining both the covariance matrix interrelating all of the parameters, and the derivative of the query response with respect to each parameter. We produced an effective system that computes these quantities, and then combines them to produce the error-bars.

The fact that our approximation is guaranteed to be correct *in the limit* does not mean it will work well in practice. We therefore empirically investigated these claims, by testing our system across a variety of different belief nets and queries, and over a range of sample sizes and credibility levels. We found that the method works well throughout.

## Acknowledgements

We are grateful for the many comments and suggestions we received from Adnan Darwiche and the anonymous reviewers, and for the fairness of the UAI'01 programme chairs. All authors greatfully acknowledge the generous support provided by NSERC, iCORE and Siemens Corporate Research. Most of this work was done while the first author was a student at the University of Alberta.

## References

[Aki96] Y. Akimoto. A note on uniform asymptotic normality of Dirichlet distribution. *Mathematica Japonica*, 44:25–30, 1996.

[BFGK96] C. Boutilier, N. Friedman, M. Goldszmidt, and D. Koller. Context-specific independence in Bayesian networks. In *UAI-96*, 1996.

[BFH95] Y. Bishop, S. Fienberg, and P. Holland. *Discrete Multivariate Analysis — Theory and Practice*. MIT Press, 1995.

[BKRK97] J. Binder, D. Koller, S. Russell, and Ke. Kanazawa. Adaptive probabilistic networks with hidden variables. *Machine Learning*, 29:213–244, 1997.

[BSCC89] I. Beinlich, H. Suermondt, R. Chavez, and G. Cooper. The ALARM monitoring system: A case study with two probabilistic inference techniques for belief networks. In *Proc. Second European Conf. Artificial Intelligence in Medicine*, August 1989.

[CH92] G. Cooper and E. Herskovits. A Bayesian method for the induction of probabilistic networks from data. *MLJ*, 9:309–347, 1992.

[CNKE93] P. Che, R. Neapolitan, J. Kenevan, and M. Evens. An implementation of a method for computing the uncertainty in inferred probabilities in belief networks. In *UAI-93*, pages 292–300, 1993.

[Coo90] G. Cooper. The computational complexity of probabilistic inference using Bayesian belief networks. *Artificial Intelligence*, 42(2–3):393–405, 1990.

[Dar00] A. Darwiche. A differential approach to inference in bayesian networks. In *UAI'00*, 2000.

[Dec98] R. Dechter. Bucket elimination: A unifying framework for probabilistic inference. In *Learning and Inference in Graphical Models*, 1998.

[GDS01] R. Greiner, C. Darken, and I. Santoso. Efficient reasoning. *Computing Surveys*, 33:1–30, 2001.

[GGS97] R. Greiner, A. Grove, and D. Schuurmans. Learning Bayesian nets that perform well. In *UAI-97*, 1997.

[HC91] E. Herskovits and C. Cooper. Algorithms for Bayesian belief-network precomputation. In *Methods of Information in Medicine*, pages 362–370, 1991.

[Hec95] March 1995. Special issue of "Communications of the ACM", on Bayesian Networks.

[Hec98] D. Heckerman. A tutorial on learning with Bayesian networks. In *Learning in Graphical Models*, 1998.

[Kle96] G. Kleiter. Propagating imprecise probabilities in bayesian networks. *Artificial Intelligence*, 88, 1996.

[Las95] K. Laskey. Sensitivity analysis for probability assessments in Bayesian networks. *IEEE Transactions on Man, Cybernetics and Systems*, 25(6):901–909, 1995.

[LS99] V. Lepar and P. P. Shenoy. A comparison of Lauritzen-Spiegelhalter, Hugin, and Shenoy-Shafer architectures for computing marginals of probability distributions. In *UAI98*, 1999.

[Mus93] R. Musick. Minimal assumption distribution propagation in belief networks. In *UAI93*, 1993.

[Pea88] J. Pearl. *Probabilistic Reasoning in Intelligent Systems: Networks of Plausible Inference*. Morgan Kaufmann, 1988.

[VGH01] T. Van Allen, R. Greiner, and P. Hooper. Bayesian error-bars for belief net inference. Technical report, University of Alberta, 2001.